# Discovering causal structures in binary exclusive-or skew acyclic models

Takanori Inazumi[1], Takashi Washio[1,4], Shohei Shimizu[1],
Joe Suzuki[2], Akihiro Yamamoto[3] and Yoshinobu Kawahara[1]

[1] The Institute of Scientific and Industrial Research, Osaka University, 8-1 Mihogaoka, Ibaraki, Osaka, Japan
[2] Graduate School of Science, Osaka University, 1-1 Machikaneyama, Toyonaka, Osaka, Japan
[3] Graduate School of Informatics, Kyoto University, 36-1 Yoshida-Honmachi, Sakyo-ku, Kyoto, Japan
[4] JST, ERATO, Minato Discrete Structure Manipulation System Project, 1-4-14 Shibata, Kita-ku, Osaka, Japan

## Abstract

Discovering causal relations among observed variables in a given data set is a main topic in studies of statistics and artificial intelligence. Recently, some techniques to discover an identifiable causal structure have been explored based on non-Gaussianity of the observed data distribution. However, most of these are limited to continuous data. In this paper, we present a novel causal model for binary data and propose a new approach to derive an identifiable causal structure governing the data based on skew Bernoulli distributions of external noise. Experimental evaluation shows excellent performance for both artificial and real world data sets.

## 1 INTRODUCTION

Many approaches to causal inference and structure learning of Bayesian networks have been studied in statistics and artificial intelligence (Pearl, 2000), (Spirtes et al., 2000). Most of these derive candidate causal structures from an observed data set by assuming acyclicity of the causal dependencies. These mainly use information up to second-order statistics in the observed variables, and narrow down the candidate directed acyclic graphs (DAGs) by using some constraints and/or scoring functions. However, these approaches often retain multiple candidate causal structures within equivalence classes and local optima.

Recently, some linear non-Gaussianity-based approaches as applied to acyclic models (LiNGAM) have been proposed (Shimizu et al., 2006). Such approaches are known to derive an identifiable causal structure of a linear structural equation model by using information on the non-Gaussian distribution of external noise driving the observed variables. However, these require linearity of the objective system and are only applicable to continuous variables. Recent studies extended this principle to non-linear models with continuous variables (Hoyer et al, 2009) and variables having ordered discrete values (Peters et al., 2011). However, these studies are not primarily applicable to more than two variables, and do not cover any variables having unordered categorical or binary values.

In contrast, many real world domains such as medicine (Pearl, 2000), bioinformatics (Shi et al., 2007) and sociology (Spirtes et al., 2000), maintain massive stochastic binary data sets, and practitioners need to discover those causal structures from the data for various purposes. However, to our best knowledge, all past approaches do not ensure an identifiable solution for any given binary data set. In this paper, our objective is to propose a new approach to discover an identifiable causal structure under some feasible assumptions within a given stochastic binary data set.

In the next section, we briefly review some related work to indicate important technical issues. In the third section, we introduce a novel binary exclusive-or skew acyclic model, termed "*BExSAM*", to represent an objective system, and characterize the model in respect to causal ordering and causal structure estimation. In the fourth, we propose novel algorithms and measures for the causal ordering and the causal structure estimation based on the model characterization. In the fifth, we present some experimental evaluations using both artificial and real world data sets.

## 2 RELATED WORK

Many studies on causal inference and structure learning of Bayesian networks have exerted much effort in developing principles for efficient searches by using information up to second-order statistics (Pearl, 2000), (Spirtes et al., 2000). That has arisen because exhaustive searches are intractable as the number of possible directed acyclic graphs (DAGs) grows exponentially with the number of variables. To address this issue, constraint-based approaches such as the PC algorithm (Spirtes et al., 2000) and score-based approaches such as the GES algorithm (Chickering, 2002) have been studied for both continuous and discrete variables. However, these admit multiple solutions within equivalence classes and local optima in many cases, and thus often fail to generate an identifiable causal structure.

A recent technique LiNGAM (Shimizu et al., 2006) formulates the DAG structure search in form of an independent component analysis (ICA), that ensures the existence of a unique global optimum under assumptions of linear relations among the observed variables and non-Gaussianity of external noises. However, it may often provide a local optimum solution because of its greedy search. In contrast, more recent DirectLiNGAM (Shimizu et al., 2009) efficiently derives an identifiable solution through its iterative findings of exogenous variables by applying linear regressions and independence measures. (Hoyer et al, 2009) and (Zhang and Hyvarinen, 2009) proposed extensions of this principle to a non-linear additive noise model and a post-nonlinear (PNL) causal model respectively. However, their identifiability has been primarily shown only for two variable cases with the exception of a few special cases. (Mooij et al., 2009) proposed a novel independence-based regression to allow causal inference in a non-linear additive noise model containing multiple variables. However, its identifiability of the causal structure is not confirmed, and its convergence to a unique solution is not ensured because of the non-convexity of the regression problem.

One main advantage of non-Gaussianity-based approaches is the capability to derive an identifiable causal structure under some feasible model assumptions. However, because non-Gaussianity is a property of continuous distributions, these are applicable to data sets containing only continuous variables. In consequence, only a few studies have addressed this issue for discrete variable sets. A study on this topic was recently reported by (Peters et al., 2011). However, its technique is applicable only to two variables having ordered discrete values, because its basic principle is essentially an extension of the aforementioned work (Hoyer et al, 2009). Another study (Sun and Janzing, 2007) proposed a principle to find a causal order of binary variables to explain a given sample distribution by mutually independent Markov kernels. However, its identifiability is not confirmed, and its applicability is limited to very simple Boolean relations because of high computational complexity of the kernel functions for generic cases.

Because binary variables do not constitute a continuous algebra, we should develop a structural model of acyclic relations among the binary variables upon other algebraic systems such as a Boolean algebra. In addition, we need to apply a binary data distribution such as a Bernoulli distribution instead of Gaussian/non-Gaussian distributions in the model. Moreover, we have to invent novel algorithms for both causal ordering and causal structure estimation based upon characteristics of the structural model and the data distribution. In the following sections, we present our ideas concerning these issues.

## 3 PROPOSED PRINCIPLES

### 3.1 BExSAM

We first introduce a novel structural model involving generic acyclic causal relations among binary variables. Our model has a DAG structure where every external noise affects its corresponding observed variable via an exclusive-or operation which is different from the continuous additive noise.

**Definition 1** *Given a set of $d$ observed binary variables $X = \{x_i | x_i \in \{0,1\}, i = 1, \ldots, d\}$. Let a set of $d$ unobserved binary noises be $E = \{e_i | e_i \in \{0,1\}, i = 1, \ldots, d\}$ where each $e_i$ is a mutually independent stochastic variable. Then, the index $i$ is a permutation $i : \{1, \ldots, d\} \to \{1, \ldots, d\}$ so that*

$$\begin{aligned}
x_{i(1)} &= f_{i(1)} \oplus e_{i(1)}, \text{ and} \\
x_{i(k)} &= f_{i(k)}(x_{i(1)}, \ldots, x_{i(k-1)}) \oplus e_{i(k)} \\
&\qquad \text{for } k = 2, \ldots, d,
\end{aligned}$$

*where $f_{i(1)}$ is a constant in $\{0,1\}$, $f_{i(k)} : \{0,1\}^{k-1} \to \{0,1\}$ is a deterministic Boolean function for $k \geq 2$, and $\oplus$ is an exclusive-or operator.*

$f_{i(k)}$ expresses any deterministic binary relation without loss of generality because such a relation is always represented by Boolean algebraic formulae (Gregg, 1998). In the rest of this paper, subscripts $i(k)$ will be shortened to $i$ for notational simplicity, unless ambiguity requires to use the full notation. In addition, $f_{i(k)}(x_{i(1)}, \ldots, x_{i(k-1)})$ is represented by $f_i$ for brevity.

Furthermore, we introduce a Bernoulli distribution and skewness to every noise as follows.

**Definition 2** *Let the probability of $e_i = 1$ be $p(e_i = 1)$. $p(e_i = 1)$ is a constant $p_i$ satisfying $0 < p_i < 1$ and $p_i \neq 1/2$ for all $i = 1, \ldots, d$.*

$p(e_i = 0) = 1 - p_i$ satisfies the same constraints $0 < p(e_i = 0) < 1$ and $p(e_i = 0) \neq 1/2$. This noise characteristic is essentially needed in our approach as an analogue to the aforementioned non-Gaussianity in the case of continuous variables. A model based on the above two definitions is called a binary exclusive-or skew acyclic model, i.e., "*BExSAM*" for short.

When a directional edge exists from a variable $x_j$ to another variable $x_i$ ($i \neq j$) in a DAG model, $x_j$ is called a "*parent*" of $x_i$ and $x_i$ is called a "*child*" of $x_j$. As widely noted in the causal inference study (Pearl, 2000), (Spirtes et al., 2000), a variable having no *parents* in a DAG model is called an "*exogenous variable*", otherwise, it is called an "*endogenous variable*." In this study, we further introduce the following definition of a particular endogenous variable.

**Definition 3** *An endogenous variable which does not have any children in a DAG model is called a "sink endogenous variable."*

This *sink endogenous variable* plays a key role in a *BExSAM* in regard to principles and algorithms for causal ordering and causal structure estimation.

**Example 1** *The following is an example of a BExSAM consisting of four binary variables with $i(k) = k$.*

$$\begin{aligned} x_1 &= e_1, \\ x_2 &= x_1 \oplus e_2, \\ x_3 &= x_1 x_2 \oplus e_3, \\ x_4 &= (x_1 + x_3) \oplus e_4. \end{aligned}$$

Here, an and operation of $x_1$ and $x_2$ and an or operation of $x_1$ and $x_3$ are represented by $x_1 x_2$ and $x_1 + x_3$ respectively for simplicity. $x_1$ is an *exogenous variable*, and $x_4$ is a *sink endogenous variable*.

### 3.2 CHARACTERIZATION

In this subsection, characteristics of *BExSAM* associated with a *sink endogenous variable* are analyzed concerning the discovery of causal structure. First, we define a notion of "*control*" which specifies the values of some observed variables in $X$.

**Definition 4** *Let $\mathcal{X}_j = \{0, 1\}$ be a domain of $x_j \in X$. Given $X_i = X \setminus \{x_i\}$, i.e., $X_i = \{x_1, \ldots, x_{i-1}, x_{i+1}, \ldots, x_d\}$, and let $V_i$ be a $(d-1)$-tuple $(v_1, \ldots, v_{i-1}, v_{i+1}, \ldots, v_d)$ where $v_j \in \mathcal{X}_j$ for all $x_j \in X_i$. We denote $X_i = V_i$ when we assign the value $v_j$ to its $x_j$ for all $x_j \in X_i$. This assignment is called a "control" of $X_i$ at $V_i$. Similarly, given $X_{ij} = X \setminus \{x_i, x_j\}$. The "control" of $X_{ij}$ at $V_{ij}$ is defined in the same way and denoted by $X_{ij} = V_{ij}$.*

We present an important theorem for causal ordering of the variables in $X$ by using the *control*.

**Theorem 1** *The following conditions are equivalent.*

1. *$x_i \in X$ is a sink endogenous variable.*

2. *Let $X_i = X \setminus \{x_i\}$. There is a common constant $q_i$ such that $p(x_i = 1 | X_i = V_i) = q_i$ or $1 - q_i$ (and therefore, $p(x_i = 0 | X_i = V_i) = 1 - q_i$ or $q_i$ equivalently) for all control $X_i = V_i \in \{0, 1\}^{d-1}$.*

*Proof.* See Appendix 1.

For example, if we are given the two *controls* $X_4 = \{x_1, x_2, x_3\}$ at $V_4 = (0, 0, 0)$ and $V_4' = (0, 0, 1)$ in Example 1, we have the following conditional probabilities of the *sink endogenous variable*: $x_4 = 1$ as

$$\begin{aligned} p(x_4 = 1 | X_4 = V_4) &= p_4, \text{ and} \\ p(x_4 = 1 | X_4 = V_4') &= 1 - p_4, \end{aligned}$$

respectively. Actually, $p(x_4 = 1 | X_4 = V_4) = p_4 (= q_4)$ or $1 - p_4 (= 1 - q_4)$ holds for any $V_4$ in this case. In contrast, if we are provided with *controls* $X_3 = \{x_1, x_2, x_4\}$ at $V_3 = (0, 0, 0)$ and $V_3' = (0, 0, 1)$,

$$p(x_3 = 1 | X_3 = V_3) = \frac{p_3 p_4}{p_3 p_4 + (1 - p_3)(1 - p_4)}, \text{ and}$$

$$p(x_3 = 1 | X_3 = V_3') = \frac{p_3 (1 - p_4)}{p_3 (1 - p_4) + (1 - p_3) p_4}$$

hold. Because $p_3, p_4 \neq 1/2$ by Definition 2, these probabilities are not equal, and also their sum is not unity. Accordingly, no constant $q_3$ or $1 - q_3$ can be assigned to both of $p(x_3 = 1 | X_3 = V_3)$ and $p(x_3 = 1 | X_3 = V_3')$ in this case. These results reflect Theorem 1, that we can find a *sink endogenous variable* in $X$ by checking the conditional probability of every variable.

Next, we present an important proposition to estimate causal structures in $X$.

**Proposition 1** *Let $x_i \in X$ be a sink endogenous variable. Given a variable $x_j \in X$ ($j \neq i$), let $X_{ij} = V_{ij}$ be a control of $X_{ij} = X \setminus \{x_i, x_j\}$ at $V_{ij} \in \{0, 1\}^{d-2}$.*

1. *$x_j$ is a parent of $x_i$. $\Leftrightarrow$ $p(x_i = v_i | x_j = v_j, X_{ij} = V_{ij}) = p(x_i = \bar{v}_i | x_j = \bar{v}_j, X_{ij} = V_{ij})$ for some $V_{ij}$,*

2. *$x_j$ is not a parent of $x_i$. $\Leftrightarrow$ $p(x_i = v_i | x_j = v_j, X_{ij} = V_{ij}) = p(x_i = v_i | x_j = \bar{v}_j, X_{ij} = V_{ij})$ for all $V_{ij}$,*

*where $v_* \in \{0, 1\}$ and $\bar{v}_* = v_* \oplus 1$.*

If $x_j$ is a parent of $x_i$, $f_i$ depends on $x_j$ under some *controls*, and its converse statement is also true. If $x_j$ is not a parent of $x_i$, $f_i$ does not depend on $x_j$ under any *controls*, and vice versa. By combining these facts with Lemma 2 in Appendix 2, we obtain this proposition. In Example 1, the *sink endogenous variable* $x_4$ satisfies the r.h.s. of the relation 1 of Proposition 1 with $x_1$ under the *control* $X_{14} = \{x_2, x_3\}$ at $V_{14} = \{0, 0\}$, and $x_1$ is a parent of $x_4$ in fact. In contrast, $x_4$ satisfies the r.h.s of the relation 2 of Proposition 1 with $x_2$, which is not a parent of $x_4$, under the *control* $X_{24} = \{x_1, x_3\}$ at $V_{24} = \{0, 0\}$. Proposition 1 indicates a way to identify parent variables of a *sink endogenous variable* by checking the conditional probability of every other variable.

## 4 PROPOSED ALGORITHMS AND INDEPENDENCE MEASURES

### 4.1 OUTLINE[1]

We propose an approach to discover an identifiable causal structure in a *BExSAM* from a given binary data set $D = \{V^{(h)}|h = 1, \ldots, n\}$ where each $V^{(h)}$ is a $d$-dimensional vector of all observed values describing $X$. This approach is based on Theorem 1 and Proposition 1. Since we assume that $D$ is generated through a process well modeled by a *BExSAM*, every element $v_i^{(h)}$ of $V^{(h)}$ is generated through $x_i = f_i \oplus e_i$ by Definition 1 where the distributions of $e_i$ are mutually independent and follow the constraints in Definition 2. Accordingly, if the sample size $n$ is sufficiently larger than $2^d$, $D$ contains varieties of vectors $V^{(h)}$, and enables to estimate the conditional probabilities under various *controls*. The required size of $D$ is tractable when the number of its observed variables is moderate as shown in numerical experiments in a later section.

Figure 1 shows the outline of our proposed algorithm. The first step computes a frequency table $FT$ of $D$. Since the algorithm uses $FT$ only in the latter steps, we can directly apply the algorithm to $FT$ if $FT$ is provided as a data set. In the loop from the next step, the algorithm seeks a *sink endogenous variable* $x_{i(k)}$ through a function "find_sink" at step 3 and the parent variables $P_{i(k)}$ of the found $x_{i(k)}$ via "find_parent" at step 4. These functions perform causal ordering and causal structure estimation in the *BExSAM*. Step 5 reduces the search space in the next loop by removing the estimated $x_{i(k)}$ from $X$ and marginalizing $x_{i(k)}$ in $FT$. The entire list of $x_{i(k)}$ and $P_{i(k)}$ in the output represents a resultant DAG structure of the *BExSAM*. This iterative reduction from the bottom in the causal order is similar to the causal ordering of (Mooij et al., 2009) while their causal structure estimation needs a second sweep from the top to the bottom.

### 4.2 CAUSAL ORDERING

Our causal ordering algorithm is summarized in Fig. 2. In the loop starting from step 1, it computes the conditional probabilities for all controls by following Theorem 1. $FT(x_i = v_i, X_i = V_i)$ at step 4 expresses the frequency of a value combination; $x_i = v_i$ and $X_i = V_i$.

---

[1] Our code is available from http://www.ar.sanken.osaka-u.ac.jp/~inazumi/bexsam.html.

---

input: a binary data set $D$ and its variable list $X$.
1. compute a frequency table $FT$ of $D$.
2. for $k := d$ to $1$ do
3.     $x_{i(k)} := $ **find_sink**$(FT, X)$.
4.     $P_{i(k)} := $ **find_parent**$(FT, X, x_{i(k)})$.
5.     remove $x_{i(k)}$ from $X$,
    and integrate $FT$ with $x_{i(k)}$.
6. end
output: a list $[\{x_{i(k)}, P_{i(k)}\}|k = 1, \ldots, d]$.

Figure 1: Main Algorithm

---

input: a frequency table $FT$ and its variable list $X$.
1. for $i := 1$ to $d$ do
2.     $X_i := X \setminus \{x_i\}$.
3.     for all $V_i \in \{0, 1\}^{d-1}$ do
4.       $p(x_i = 1|X_i = V_i) :=$
$\frac{FT(x_i=1,X_i=V_i)}{FT(x_i=0,X_i=V_i)+FT(x_i=1,X_i=V_i)}$.
5.     end
6.     compute an independence score $S_B(x_i)$ from $FT$ and $p(x_i = 1|X_i = V_i)$ for all $V_i \in \{0, 1\}^{d-1}$.
7. end
8. select $x_i$ having the best score $S_B(x_i)$ in $X$.
output: $x_i$.

Figure 2: Algorithm of find_sink

At step 6, an independence score $S_B(x_i)$ of each variable $x_i$ is computed. Finally, the variable $x_i$ having the best score, *i.e.*, the best fit to condition 2 in Theorem 1, is selected as a sink endogenous variable.

When $x_i$ is a sink endogenous variable, $p(x_i = 1|X_i = V_i)$ takes one of the values $q_i$ or $1 - q_i$ depending on the controls. However, its "entropy" $H = -q_i \log q_i - (1 - q_i) \log(1 - q_i)$ is independent of the specific control, because the entropy is invariant under the interchange of $q_i$ and $1 - q_i$. Accordingly, we define the following variance of conditional entropy values $H(x_i|X_i = V_i)$ of $p(x_i = 1|X_i = V_i)$ over the controls as $S_B(x_i)$.

$$S_B(x_i) = E[(H(x_i|X_i = V_i) - E[H(x_i|X_i = V_i)])^2],$$

where the expectation $E$ is taken over $FT$ by weighting every $H(x_i|X_i = V_i)$ with the associated frequency of $X_i = V_i$ in $FT$. The minimum value of $S_B(x_i)$ is the best, because this score is zero if the conditional entropy of $x_i$ is perfectly independent of the control.

### 4.3 STRUCTURE ESTIMATION

Figure 3 outlines our algorithm to estimate a causal structure. In the loop beginning from step 2, the conditional probability for each *control* is computed by following Proposition 1. $FT(x_i = v_i, x_j = v_j, X_{ij} = V_{ij})$

---
input: a frequency table $FT$, its variable list $X$
and a *sink endogenous variable* $x_i$.
1. $P_i = \phi$.
2. for $j := 1$ to $d$ and $j \neq i$ do
3.    $X_{ij} := X \setminus \{x_i, x_j\}$.
4.    for all $V_{ij} \in \{0,1\}^{d-2}$ do
5.      $p(x_i = 1 | x_j = v_j, X_{ij} = V_{ij}) := \frac{FT(x_i=1, x_j=v_j, X_{ij}=V_{ij})}{FT(x_i=0, x_j=v_j, X_{ij}=V_{ij}) + FT(x_i=1, x_j=v_j, X_{ij}=V_{ij})}$
     for $v_j \in \{0, 1\}$.
6.      compute an independence score $S_P(x_j, V_{ij})$ from $FT$ and $p(x_i = 1 | x_j = v_j, X_{ij} = V_{ij})$ for $v_j \in \{0, 1\}$.
7.    end
8.    apply multiple comparison tests over all $S_P(x_j, V_{ij})$ ($V_{ij} \in \{0,1\}^{d-2}$) to judge if $x_j$ is a parent of $x_i$. Include the parent $x_j$ into $P_i$.
9. end
output: $P_i$.
---

Figure 3: Algorithm of find_parent

at step 5 is the frequency of a value combination; $x_i = v_i$, $x_j = v_j$ and $X_{ij} = V_{ij}$. Furthermore, an independence score $S_P(x_j, V_{ij})$ of each variable $x_j$ is computed at step 6. Finally, variables $x_j$ having low $S_P(x_j, V_{ij})$ are selected as parents of a *sink endogenous variable* $x_i$ through multiple comparison tests.

The frequency $FT(x_i = 1, x_j = v_j, X_{ij} = V_{ij})$ which is required in the numerator at step 5 follows a binominal distribution $B(FT(x_j = v_j, X_{ij} = V_{ij}), p(x_i = 1 | x_j = v_j, X_{ij} = V_{ij}))$. Thus, we easily derive the variance $\sigma^2(X_{ij} = V_{ij})$ of $\delta p(X_{ij} = V_{ij}) := p(x_i = 1 | x_j = v_j, X_{ij} = V_{ij}) - p(x_i = 1 | x_j = \bar{v}_j, X_{ij} = V_{ij})$ and compute $S_P(x_j, V_{ij})$ which is a P-value of null-hypothesis: $p(x_i = 1 | x_j = v_j, X_{ij} = V_{ij}) = p(x_i = 1 | x_j = \bar{v}_j, X_{ij} = V_{ij})$ by using Gaussian approximation. We test this null-hypothesis over all *controls* $V_{ij} \in \{0,1\}^{d-2}$ by multiple comparison tests (Benjamini, 1995). If accepted, we cannot say that condition 2 in Proposition 1 does not hold, and we presume that $x_j$ is not a parent of $x_i$. Otherwise, condition 1 in Proposition 1 is likely to hold, and $x_j$ is judged to be a parent. These multiple comparison tests do not cause severe failures in the causal structure estimation, because their errors are localized into the parent estimation of each *sink endogenous variable*. This is different from the errors of multiple conditioning used in PC algorithms which can induce global failures. We should note that a significance level $\alpha$ used in these tests is a unique parameter in our algorithms.

### 4.4 COMPUTATIONAL COMPLEXITY

The largest table used in the above algorithms is the frequency table $FT$ of size $2^d$. According to the requirement on data size, $n > 2^d$, the memory complexity given by the size of $FT$ is $O(n)$.

The loop involved in the "find_sink" function compute the conditional probabilities $2^{d-1}$ times, and compute a score $S_B(x_i)$ by aggregating these $2^{d-1}$ probabilities and the frequency table $FT$. Thus, these are $O(n)$. Since the loop repeats $d$ times, the time complexity of "find_sink" is $O(dn)$. Similarly, the loop involved in the "find_parent" function computes the conditional probabilities and scores $2^{d-2}$ times. The multiple comparison tests requires sorting $2^{d-2}$ scores. Therefore, the time complexity is $O(n)$. Because these are repeated $d$ times in the outer loop, "find_parent" is $O(dn)$. Step 1 of the main algorithm needs $n$ counts only. The functions "find_sink" and "find_parent" which are $O(dn)$ are repeated $d$ times in the main algorithm. Accordingly, the total time complexity of the proposed algorithms is $O(d^2n)$. This complexity is favorable compared with past work. For example, DirectLiNGAM which also has an iterative algorithm structure and is considered to be one of the most efficient algorithms has $O(d^3n)$ complexity.

## 5 EXPERIMENTAL EVALUATION

### 5.1 BASIC PERFORMANCE FOR ARTIFICIAL DATA

For numerical experiments, we generated artificial data by using *BExSAMs* where every $f_{i(k)}$ is represented by the following algebraic normal form (ANF).

$$\begin{aligned}
f_{i(k)} =\ & a_0^k \oplus \\
& a_1^k x_{i(1)} \oplus a_2^k x_{i(2)} \oplus \ldots \oplus a_d^k x_{i(d)} \oplus \\
& a_{1,2}^k x_{i(1)} x_{i(2)} \oplus \ldots \oplus a_{k-1,k}^k x_{i(k-1)} x_{i(k)} \oplus \\
& \ldots \oplus \\
& a_{1,2,\ldots,k}^k x_{i(1)} x_{i(2)} \ldots x_{i(k)} \oplus e_{i(k)}.
\end{aligned}$$

Each coefficient $a_*^k$ is a constant in $\{0, 1\}$. Under given parameters of $d$, $n$ and $p(a_*^k = 1)$, we randomly determine the permutation $i : \{1, \ldots, d\} \to \{1, \ldots, d\}$ in Definition 1, and randomly generate all $a_*^k$ in this expression for all $k = 1, \ldots, d$. Because any Boolean formula $f_{i(k)}$ is known to be represented by an ANF with arbitrary coefficients $a_*^k$ (Carlet and Guillot, 1999), a set of the randomized $a_*^k$ generates a generic *BExSAM*. Moreover, we randomly generate $e_{i(1)}^{(h)}, \ldots, e_{i(d)}^{(h)}$ ($h = 1, \ldots, n$) under given $p(e_{i(k)} = 1)$ ($k = 1, \ldots, d$), and obtain $D = \{V^{(h)} | h = 1, \ldots, n\}$ through the generated *BExSAM*. Similarly to the other constraint based approaches (Spirtes et al., 2000), our algorithm basically requires a complete data set $D$ where all value patterns of $X$ are included as mentioned in subsection 4.1. Accordingly, any incomplete $D$ is excluded if it is generated. A series of the model generation, the data generation and our approach application is repeated 1000 times.

Three performance indices are used for the evaluation. Given an adjacency matrix $B$ representing the true DAG structure in a *BExSAM* which we generated. According to the estimated order of the variables in the output of Main Algorithm in Fig. 1; $[x_{i(k)}, k = 1, \ldots, d]$, the rows and the columns of $B$ are simultaneously permuted. Then, we compute an error rate $0 \leq ER_o \leq 1$ defined by a rate of nonzero elements in the strict upper triangular part of the permuted $B$, because the true $B$ is strictly lower triangular. This $ER_o$ was also used in (Shimizu et al., 2006) to evaluate the accuracy of the causal ordering. Another index is an error rate $0 \leq ER_s \leq 1$ defined by a rate of mismatched elements in the estimation $\hat{B}$ against its true $B$. $\hat{B}$ is derived from the output of Main Algorithm in Fig. 1; $[\{x_{i(k)}, P_{i(k)}\}|k = 1, \ldots, d]$. $ER_s$ reflects the accuracy of the causal structure estimation together with the given causal order. The third is simply a total computation time $CT$ (msec) of our algorithms explained in section 4. These indices are averaged over the 1000 trials.

In the first experiment, every combination of $d = 2, 4, 6, 8$ and $n = 100, 500, 1000, 5000, 10000$ was evaluated with the probabilities $0 \leq p(a_*^k = 1) \leq 1$ and $p(e_{i(k)} = 1) \in \{0.1, 0.2, 0.3, 0.4, 0.6, 0.7, 0.8, 0.9\}$ chosen at uniformly random for all $k = 1, \ldots, d$. This choice of $p(e_{i(k)} = 1)$ ensures the skew and non-deterministic distribution of each $e_{i(k)}$ requested by Definition 2. $p(a_*^k = 1)$ reflects density of the variable couplings in the generated *BExSAM*. Table 1 summarizes the performance of our approach. The experiments for $n = 100, 500$ and $d = 8$ were skipped, since $n$ were insufficient for the completeness of $D$. Both error rates $ER_o$ and $ER_s$ are less than or around 0.05 if $n \geq 1000$ and $d \leq 4$. This indicates that our approach works for a moderate data size around 1000 if we can limit the number of variables less than or equal to 4. For the case of $d \geq 6$, we need to collect a large number of data more than 5000. $CT$ particularly increases along with $d$. This is consistent with the aforementioned complexity analysis, but does not cause practical problems since the required $CT$ is very small even for $d = 8$ and $n = 10000$.

In the second experiment, we gave an identical value to every $p(e_{i(k)} = 1)$ of all $k = 1, \ldots, d$, and evaluated $ER_o$ and $ER_s$ under $d = 4$, $n = 1000$ and the aforementioned random $p(a_*^k = 1)$. Figure 4 depicts the resultant relations of $ER_o$ and $ER_s$ with $p(e_{i(k)} = 1)$. If $p(e_{i(k)} = 1)$ is close to 0 or 1, our approach loses the chance to accurately estimate the conditional probabilities and thus its accuracy is degraded. If $p(e_{i(k)} = 1)$ is closed to 0.5, again the accuracy of our approach is lost, because our approach totally relies on Theorem 1 and Proposition 1 which require $p(e_{i(k)} = 1) \neq 0.5$. Through some extra experiments, we confirmed that $ER_o$ and $ER_s$ do not show strong dependency on $p(a_*^k = 1)$, i.e., the density of the *BExSAM*. We also confirmed that a false negative rate to miss true parent variables is dominant rather than a false positive rate to find spurious parent variables. This is reasonable since the multiple comparison tests check a null hypothesis that a variable is not a parent. The experiments also showed that the computation time $CT$ is almost independent of both $p(a_*^k = 1)$ and $p(e_{i(k)} = 1)$.

Table 1: Performance under various $n$ and $d$ (top:$ER_o$, middle:$ER_s$ and bottom:$CT$ (msec) in each cell)

| $n \backslash d$ | 2 | 4 | 6 | 8 |
|---|---|---|---|---|
| | 0.035 | 0.20 | 0.29 | - |
| 100 | 0.043 | 0.20 | 0.34 | - |
| | 0.96 | 2.6 | 7.1 | - |
| | 0.004 | 0.033 | 0.23 | - |
| 500 | 0.0095 | 0.068 | 0.24 | - |
| | 0.90 | 2.7 | 7.9 | - |
| | 0.00 | 0.017 | 0.16 | 0.15 |
| 1000 | 0.0043 | 0.041 | 0.18 | 0.25 |
| | 0.89 | 2.8 | 8.3 | 32.0 |
| | 0.00 | 0.0015 | 0.042 | 0.16 |
| 5000 | 0.0063 | 0.011 | 0.055 | 0.16 |
| | 1.1 | 3.4 | 9.9 | 35.0 |
| | 0.00 | 0.0013 | 0.020 | 0.14 |
| 10000 | 0.0050 | 0.0076 | 0.027 | 0.14 |
| | 1.2 | 4.2 | 11.0 | 40.0 |

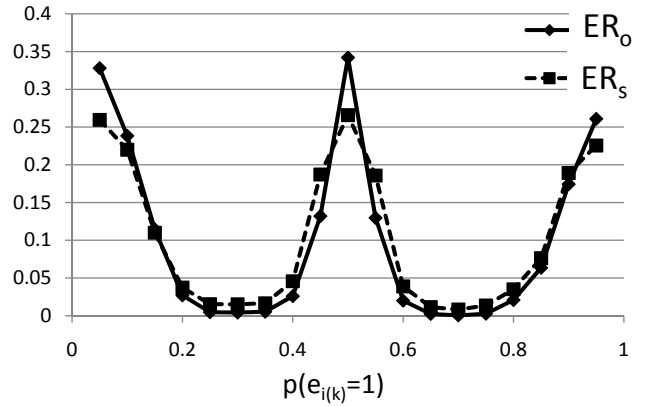

Figure 4: Dependency of $ER_o$ and $ER_s$ on $p(e_{i(k)} = 1)$

## 5.2 COMPARISON WITH OTHER ALGORITHM

Our algorithm, PC algorithm (Spirtes et al., 2000), CPC algorithm (Ramsey et al., 2006) and GES algorithm (Chickering, 2002) were compared through their applications to data generated by a following artificial *BExSAM* forming a Y structure (Mani et al., 2006).

Table 2: Performance of two algorithms

| Our Algorithm | | | |
|---|---|---|---|
| **true** \ **est.** | directed | no edge | undirected |
| directed | *53* | 7 | 0 |
| no edge | 7 | *173* | 0 |
| PC Algorithm | | | |
| **true** \ **est.** | directed | no edge | undirected |
| directed | *31* | 20 | 9 |
| no edge | 9 | *162* | 9 |
| CPC Algorithm | | | |
| **true** \ **est.** | directed | no edge | undirected |
| directed | *32* | 13 | 15 |
| no edge | 1 | *162* | 17 |
| GES Algorithm | | | |
| **true** \ **est.** | directed | no edge | undirected |
| directed | *33* | 13 | 14 |
| no edge | 0 | *164* | 16 |

$$\begin{aligned} x_1 &= e_1 \\ x_2 &= e_2 \\ x_3 &= x_1 x_2 \oplus e_3 \\ x_4 &= x_3 \oplus e_4 \end{aligned}$$

Each $p(e_{i(k)} = 1)$ was given similarly to the first experiment in the previous subsection. Table 2 shows frequencies of estimated relations between variables corresponding to its columns over their true relations corresponding to its row for 20 trials. Because of the Y structure among four variables, the number of the true directed edges is $3 \times 20$ trials $= 60$ in total while the number of the rest true no edges are $(4 \times 3 - 3) \times 20$ trials $= 180$ by double counting two missed directed edges between two variables for a no edge. This counting method gives double penalties to a wrong estimation of an edge direction which often comes from causal ordering failures affecting the global structure estimation. The *italics* show numbers of the correct estimations. Note that this Y structure is a typical example which enables a valid estimation of the PC algorithm. However, our approach based on the skewness of the binary data distribution provides better accuracy. Similar advantageous results of our approach was obtained for the case where $x_3 = (x_1 + x_2) \oplus e_3$. The results of CPC and GES are similar with PC, since CPC and GES do not have any significant advantage to identify Y-structures comparing with PC.

### 5.3 EXAMPLE APPLICATIONS TO REAL-WORLD DATA

Our approach has been applied to two real-world data sets. One is on leukemia deaths/survivals ($LE = 1/0$) in children in southern Utah with high/low exposure to radiation ($EX = 1/0$) from the fallout of nuclear tests in Nevada (Finkelstein and Levin, 1990), (Pearl, 2000). As this contains only two binary variables, conventional constraints/score-based approaches cannot estimate any unique causal structure. In contrast, our approach found a causal order $EX \rightarrow LE$ consistent with our intuition, though their parent-child relation was not significant under $\alpha = 0.05$ in the multiple comparison tests.

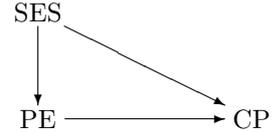

Figure 5: Discovered causal structure among $CP$, $PE$ and $SES$

Another data set is on college plans of 10318 Wisconsin high school seniors (Sewell and Shah, 1968), (Spirtes et al., 2000). While the original study aimed to find a feasible causal structure among five variables constituting a Y structure, we focus on the causality between three variables; yes/no college plans ($CP = 0/1$), low/high parental encouragement ($PE = 0/1$) and lowest to highest socioeconomic status ($SES = 0, \ldots, 3$). Conventional constraints-based approaches are known to give multiple candidate causal structures in an equivalence class for the three variables. We selected 3756 male seniors ($SEX = male$) having a higher intelligence quotient by excluding the lowest group ($IQ \neq lowest$) to *control* their individual characters while maintaining the sample size. We further transformed $SES$ to a binary variable according to the observation that the samples having its lowest value have a particular low probability of yes college plans; $p(CP = 0) = 0.21$ while the other samples have $p(CP = 0) = 0.51$. To preserve this strong dependency, it is discretized as $SES = 0$ if it is originally the lowest, and $SES = 1$ otherwise. The application of our approach with a significance level $\alpha = 0.05$ to this data set provided a unique causal structure depicted in Fig. 5. It stats that the socioeconomic status of the parents affects both the parental encouragement and the college plan of a senior, and the parental encouragement further influences the college plan. This is consistent with our intuition and the structure estimated from the original five variables constituting a Y structure by the PC algorithm.

## 6 DISCUSSION

When we apply our approach to a data set, we presume that the data generation process approximately follows a *BExSAM*. A crucial property requested in a *BExSAM* is the skew distribution of every external noise. Because the noise is not observable, a measure to indirectly check this property by using the given observed data set is desirable. The following lemma can be used for this purpose.

**Lemma 1** *Assuming that a data set $D$ is generated by a BExSAM, if a variable $x_i \in X$ has a distribution $p(x_i = 1) \neq 0.5$, then the following conditions hold.*

$$p(e_i = 1) \neq 0.5 \text{ and } p(f_i = 1) \neq 0.5,$$

where $x_i = f_i \oplus e_i$.
Proof. See Appendix 3.

We simply check if the frequency of $x_i = 1$ is apart from 0.5 for every observed variable in $X$. If it is, the skewness of their noise distribution is ensured. Another strong assumption of a *BExSAM* is the interventions of external binary noises via exclusive-or operations. However, this Lemma 1 also suggests the applicability of the *BExSAM* to generic Boolean interventions of the noises. For example, if a given data is generated by $x_i = f_i + e_i$ (+ is an or operator.), we have $p(x_i = 1) = p(e_i = 1) + p(f_i = 1) - p(e_i = 1)p(f_i = 1)$. On the other hand, a *BExSAM* $x_i = f_i \oplus e_i$ provides $p(x_i = 1) = p(e_i = 1) + p(f_i = 1) - 2p(e_i = 1)p(f_i = 1)$. Accordingly, these two models show very similar distributions of the observed variables, if $p(e_i = 1), p(f_i = 1) \ll 0.5$. These conditions can be checked by the insights of Lemma 1.

As mentioned in subsection 5.1, our algorithm basically requires a complete data set $D$ similarly to the other constraint based approaches (Spirtes et al., 2000). Therefore, to analyze an incomplete data set $D$, we need to estimate the missing data in $D$ by introducing some data completion techniques such as (Bernaards et al., 2007). Another associated issue is its applicability to many variables, since the low error rates are ensured only for less than 10 variables with thousands of samples. A promising way to overcome this issue may be to combine our approach with the other constraint-based approaches such as the PC algorithm as discussed in (Zhang and Hyvarinen, 2009). The extensions of our approach toward these issues are remained for future studies.

## 7 CONCLUSION

In this paper, we presented a novel binary structural model involving exclusive-or noise and proposed a new efficient approach to derive an identifiable causal structure governing a given binary data set based on the skewness of the distributions of external noises. The approach has low computational complexity and requires only one tunable parameters. The experimental evaluation shows promising performance for both artificial and real world data sets.

This study provides an extension of the non-Gaussianity-based causal inference for continuous variables to the causal inference for discrete variables, and suggests a new aspect on more generic causal inference.

**Acknowledgements**

This work was partially supported by JST, ERATO, Minato Discrete Structure Manipulation System Project and JSPS Grant-in-Aid for Scientific Research(B) #22300054. The authors would like to thank Dr. Tsuyoshi Ueno, a research fellow of the JST ERATO project, for his valuable technical comments.

**Appendix 1**

**Proof of Theorem 1**
($1 \Rightarrow 2$)
Let the value of $x_i$ be $v_i \in \{0,1\}$. Under a control $X_i = V_i$, $f_i$ is a constant in $\{0,1\}$. Accordingly, the following relation holds by Definition 1.

$$x_i = v_i \Leftrightarrow v_i = f_i \oplus e_i \Leftrightarrow e_i = v_i \oplus f_i.$$

Because $x_i$ is a sink endogenous variable, $e_i$ and $X_i$ are mutually independent by Definition 1. Under this fact and Definition 2,

$$p(x_i = v_i | X_i = V_i) = p(e_i = v_i \oplus f_i)$$
$$= \begin{cases} p_i, & \text{for } v_i \oplus f_i = 1 \\ 1 - p_i, & \text{for } v_i \oplus f_i = 0 \end{cases}$$

By letting $q_i = p_i$ or $q_i = 1 - p_i$, $1 \Rightarrow 2$ holds.
($1 \Leftarrow 2$)
Assume that $x_i$ is not a sink endogenous variable. Let $X_i$ be partitioned into $X_i^l$ and $X_i^u$ where $X_i^l$ is a set of all descendants of $x_i$ in a BExSAM, and $X_i^u$ is the complement of $X_i^l$ in $X_i$. Then, the following holds.

$$p(x_i = v_i | X_i = V_i) = p(x_i = v_i | X_i^u = V_i^u, X_i^l = V_i^l)$$
$$= \frac{p(X_i^l = V_i^l | x_i = v_i, X_i^u = V_i^u) p(x_i = v_i | X_i^u = V_i^u)}{\sum_{v' \in \{v_i, \bar{v}_i\}} p(X_i^l = V_i^l | x_i = v', X_i^u = V_i^u) p(x_i = v' | X_i^u = V_i^u)}, \quad (t1).$$

where $\bar{v}_i = v_i \oplus 1$. Furthermore, let $X_j = X \setminus \{x_j\}$ of $x_j \in X_i^l$ be partitioned into $X_j^l$ and $X_j^u$ similarly to $X_i^l$ and $X_i^u$ of $x_i$. By Definition 1, each $f_j$ of $x_j \in X_i^l$ is given by $X_j^u = V_j^u$, and thus $x_j = f_j(X_j^u = V_j^u) \oplus e_j$ for all $x_j \in X_i^l$. Accordingly, $X_i^l = V_i^l$ is equivalent to $e_j = v_j \oplus f_j(X_j^u = V_j^u)$ for all $x_j \in X_i^l$. We rewrite the r.h.s.: $v_j \oplus f_j(X_j^u = V_j^u)$ as $l_j(v_i)$, since $x_i$ is an ancestor of $x_j$, i.e., $x_i \in X_j^u$, and the values of all variables except $x_i = v_i$ are constant under a control $X_i = V_i$. Because every $e_j$ is independent of its upper variables,

$$p(X_i^l = V_i^l | x_i = v_i, X_i^u = V_i^u) = \prod_{x_j \in X_i^l} p(e_j = l_j(v_i)).$$

$x_i$ has at least one child $x_h \in X_i^l$ where $l_h(1) \neq l_h(0)$ for some control $X_i = V_i$ from the assumption that $x_i$ is not a sink endogenous variable. Accordingly,

$$p(X_i^l = V_i^l | x_i = v_i, X_i^u = V_i^u) \quad (t2)$$
$$= \begin{cases} p_h \prod_{x_j \in X_i^l, x_j \neq x_h} p(e_j = l_j(v_i)), & \text{for } l_h(v_i) = 1 \\ (1-p_h) \prod_{x_j \in X_i^l, x_j \neq x_h} p(e_j = l_j(v_i)), & \text{for } l_h(v_i) = 0 \end{cases}.$$

On the other hand, since $e_i$ is independent of $X_i^u$ and $x_i = v_i \leftrightarrow e_i = v_i \oplus f_i$,

$$p(x_i = v_i | X_i^u = V_i^u) = p(e_i = v_i \oplus f_i) \quad (t3)$$
$$= \begin{cases} p_i, & \text{for } v_i \oplus f_i = 1 \\ 1 - p_i, & \text{for } v_i \oplus f_i = 0 \end{cases}.$$

By substituting Eq.($t2$) and ($t3$) to Eq.($t1$), we obtain the following four cases.

$$p(x_i = v_i | X_i = V_i) =$$

$$\frac{\alpha p_i p_h}{\alpha p_i p_h + \beta (1-p_i)(1-p_h)} \quad (t4), \quad \frac{\alpha p_i (1-p_h)}{\alpha p_i (1-p_h) + \beta (1-p_i) p_h} \quad (t5),$$
$$\frac{\alpha (1-p_i) p_h}{\alpha (1-p_i) p_h + \beta p_i (1-p_h)} \quad (t6), \quad \frac{\alpha (1-p_i)(1-p_h)}{\alpha (1-p_i)(1-p_h) + \beta p_i p_h} \quad (t7),$$

where $\alpha = \prod_{x_j \in X_i^l, x_j \neq x_h} p(e_j = l_j(v_i))$ and $\beta = \prod_{x_j \in X_i^l, x_j \neq x_h} p(e_j = l_j(\bar{v}_i))$. $\alpha$ and $\beta$ are nonzero

from Definition 2, and each of (t4) = (t5), (t4) = (t6), (t5) = (t7) and (t6) = (t7) for any $\alpha$ and $\beta$, i.e., any $X_i^l = V_i^l$ excluding $x_h$, implies $p_i = 1/2$ or $p_h = 1/2$ respectively. Because neither of $p_i \neq 1/2$ nor $p_h \neq 1/2$ are allowed by Definition 2, this implies that all conditions (t4) $\neq$ (t5), (t4) $\neq$ (t6), (t5) $\neq$ (t7) and (t6) $\neq$ (t7) hold simultaneously under some $X_i^l = V_i^l$ excluding $x_h$. If we assume (t4) = (t7) and (t5) = (t6) simultaneously under such $X_i^l = V_i^l$ excluding $x_h$, they imply $p_i^2 p_h^2 = (1-p_i)^2(1-p_h)^2$ and $p_i^2(1-p_h)^2 = (1-p_i)^2 p_h^2$, i.e., $p_i = 1/2$ and $p_h = 1/2$. Accordingly, $p_i \neq 1/2$ and $p_h \neq 1/2$ from Definition 2 imply that one of (t4) = (t7) and (t5) = (t6) do not hold under the $X_i^l = V_i^l$ excluding $x_h$. This result shows that $p(x_i = v_i | X_i = V_i)$ takes more than two values for some given control $X_i = V_i$ if $x_i$ is not a sink endogenous variable. By taking its contrapositive, we obtain 1 $\Leftarrow$ 2. ∎

## Appendix 2

**Lemma 2** Let $x_i \in X$ be a sink endogenous variable. For a variable $x_j \in X$ ($j \neq i$), the followings hold under a control of $X_{ij} = X \setminus \{x_i, x_j\}$ at $V_{ij} \in \{0,1\}^{d-2}$.

1. $f_i$ depends on $x_j$ under $X_{ij} = V_{ij}$. $\Leftrightarrow$ $p(x_i = v_i | x_j = v_j, X_{ij} = V_{ij}) = p(x_i = \bar{v}_i | x_j = \bar{v}_j, X_{ij} = V_{ij})$,

2. $f_i$ does not depend on $x_j$ under $X_{ij} = V_{ij}$. $\Leftrightarrow$ $p(x_i = v_i | x_j = v_j, X_{ij} = V_{ij}) = p(x_i = v_i | x_j = \bar{v}_j, X_{ij} = V_{ij})$,

where $v_* \in \{0,1\}$ and $\bar{v}_* = v_* \oplus 1$.

*Proof.* $x_i = f_i \oplus e_i$ from Definition 1, in addition, $e_i$ and $X_{ij} = V_{ij}$ are mutually independent since $x_i$ is a sink endogenous variable. Accordingly,

$$p(x_i = v_i | x_j = v_j, X_{ij} = V_{ij})$$
$$= (1-p_i) p(f_i = v_i | x_j = v_j, X_{ij} = V_{ij}) \quad (l1)$$
$$+ p_i p(f_i = \bar{v}_i | x_j = v_j, X_{ij} = V_{ij}).$$

Since $f_i$ is a deterministic Boolean function, it is always represented by a Boolean algebraic formula. If $f_i$ depends on $x_j$ under $X_{ij} = V_{ij}$, $f_i = c \oplus x_j$ where $c$ is a constant in $\{0,1\}$. From Eq.(l1),

$$p(x_i = v_i | x_j = v_j, X_{ij} = V_{ij})$$
$$= (1-p_i) p(c \oplus v_j = v_i | x_j = v_j, X_{ij} = V_{ij})$$
$$+ p_i p(c \oplus v_j = \bar{v}_i | x_j = v_j, X_{ij} = V_{ij}).$$
$$= \begin{cases} 1 - p_i, & \text{for } c \oplus v_j = v_i \Leftrightarrow c \oplus \bar{v}_j = \bar{v}_i \\ p_i, & \text{for } c \oplus v_j = \bar{v}_i \Leftrightarrow c \oplus \bar{v}_j = v_i \end{cases}$$

Therefore,

$$p(x_i = v_i | x_j = v_j, X_{ij} = V_{ij}) =$$
$$p(x_i = \bar{v}_i | x_j = \bar{v}_j, X_{ij} = V_{ij}) \quad (l2)$$

holds. If $f_i$ does not depend on $x_j$ under $X_{ij} = V_{ij}$, $f_i = c$ where $c$ is a constant in $\{0,1\}$. From Eq.(l1),

$$p(x_i = v_i | x_j = v_j, X_{ij} = V_{ij})$$
$$= (1-p_i) p(c = v_i | x_j = v_j, X_{ij} = V_{ij})$$
$$+ p_i p(c = \bar{v}_i | x_j = v_j, X_{ij} = V_{ij}).$$
$$= \begin{cases} 1 - p_i, & \text{for } c = v_i \\ p_i, & \text{for } c = \bar{v}_i \end{cases}$$

Therefore, $p(x_i = v_i | x_j = v_j, X_{ij} = V_{ij})$ does not depend on $v_j$, and thus

$$p(x_i = v_i | x_j = v_j, X_{ij} = V_{ij}) =$$
$$p(x_i = v_i | x_j = \bar{v}_j, X_{ij} = V_{ij}) \quad (l3)$$

holds. $f_i = c \oplus x_j$ and $f_i = c$ are mutually exclusive, and one or the other holds. Eq.(l2) and Eq.(l3) are also mutually exclusive since $p_i \neq 1/2$ from Definition 2, and only one of these holds. Accordingly, this lemma holds. ∎

## Appendix 3

**Proof of Lemma 1**

Without loss of generality, $p(e_i = 1)$, $p(e_i = 0)$, $p(f_i = 1)$, $p(f_i = 0)$ are represented as

$$p(e_i = 1) = \frac{1 + \epsilon_i}{2}, \; p(e_i = 0) = \frac{1 - \epsilon_i}{2} \quad (-1 \leq \epsilon_i \leq 1),$$

$$p(f_i = 1) = \frac{1 + \xi_i}{2}, \; p(f_i = 0) = \frac{1 - \xi_i}{2} \quad (-1 \leq \xi_i \leq 1).$$

Because $e_i$ and $f_i$ are mutually independent, and have a relation $x_i = f_i \oplus e_i$,

$$\begin{aligned} p(x_i = 1) &= p(e_i = 1)p(f_i = 0) + p(e_i = 0)p(f_i = 1) \\ &= \frac{(1+\epsilon_i)(1-\xi_i)}{4} + \frac{(1-\epsilon_i)(1+\xi_i)}{4} \\ &= \frac{1 - \epsilon_i \xi_i}{2}, \\ p(x_i = 0) &= p(e_i = 1)p(f_i = 1) + p(e_i = 0)p(f_i = 0) \\ &= \frac{(1+\epsilon_i)(1+\xi_i)}{4} + \frac{(1-\epsilon_i)(1-\xi_i)}{4} \\ &= \frac{1 + \epsilon_i \xi_i}{2}. \end{aligned}$$

$\Rightarrow p(x_i = 0) - p(x_i = 1) = \epsilon_i \xi_i$.

Accordingly, if $p(x_i = 1) \neq p(x_i = 0)$ then $\epsilon_i \xi_i \neq 0$. This implies $\epsilon_i \neq 0$ and $\xi_i \neq 0$. Therefore, if $p(x_i = 1) \neq 0.5$ then $p(e_i = 1) \neq 0.5$ and $p(f_i = 1) \neq 0.5$ because of $p(x_i = 0) = 1 - p(x_i = 1)$, $p(e_i = 0) = 1 - p(e_i = 1)$ and $p(f_i = 0) = 1 - p(f_i = 1)$. ∎